\documentclass{esannV2}
\usepackage[dvips]{graphicx}
\usepackage[latin1]{inputenc}
\usepackage{amssymb,amsmath,array}
\usepackage{amsthm}
\usepackage{caption}

\usepackage{listings}
\usepackage{ifxetex}
\usepackage{fancyhdr}
\usepackage{hyperref}
\usepackage{graphicx}
\usepackage{wrapfig}
\usepackage{amsmath}
\usepackage{amsfonts}
\usepackage{amssymb}
\usepackage{listings}
\usepackage{setspace}
\usepackage{wrapfig}
\usepackage{tablefootnote}
\usepackage{multirow}
\usepackage{booktabs}
\usepackage[section]{placeins}
\usepackage{color}
\usepackage{amsfonts}
\usepackage{ulem}
\usepackage{algorithm}
\usepackage{algorithmicx}
\usepackage{algpseudocode}
\usepackage{graphicx}
\usepackage{float}
\usepackage{dsfont}
\usepackage{cleveref}

\newcommand{\set}[1]{\mathcal{#1}}

\DeclareMathOperator*{\loss}{{\ell}}

\DeclareMathOperator*{\argmax}{\arg\max}

\newcommand{\RN}{\mathbb{R}}

\newcommand{\lZero}{l_0}
\usepackage{footnote}
\newcommand{\refdef}[1]{Definition~\ref{#1}}

\newcommand{\refeq}[1]{Eq.~\eqref{#1}}

\DeclareMathOperator*{\regularization}{\ensuremath{{\theta}}}
\newcommand{\x}{\ensuremath{\vec{x}}}

\newcommand{\y}{\ensuremath{y}}

\newcommand{\xorig}{\ensuremath{\vec{x}_{\text{orig}}}}

\newcommand{\setX}{\ensuremath{\set{X}}}
\newcommand{\setY}{\ensuremath{\set{Y}}}

\newcommand{\xcf}{\ensuremath{\vec{x}_{\text{cf}}}}

\newcommand{\ycf}{\ensuremath{y'}}

\newcommand{\dimsym}{d}

\newcommand{\classifier}{\ensuremath{h}}

\newcommand{\reject}{\ensuremath{r}}
\newcommand{\threshold}{\ensuremath{\theta}}

\newcommand{\myCF}[2]{\ensuremath{\text{CF}(#1,#2)}}
\newcommand{\rejectSymbol}{\infty}
\newcommand{\xnew}{\ensuremath{\x_{*}}}

\newcommand{\Dcalib}{\ensuremath{\set{D}_\text{calib}}}
\newcommand{\Dlocal}{\ensuremath{\set{D}_\text{local}}}
\newcommand{\classifierLocal}{\ensuremath{\classifier_{\text{local}}}}
\newcommand{\nonconformity}{\ensuremath{\phi}}
\newcommand{\credibility}{\ensuremath{\psi}}

\newcommand{\explanation}{\ensuremath{\Psi}}
\newcommand{\featurerelvance}{\ensuremath{\text{FRI}}}

\usepackage{amsthm}

\newtheorem{definition}{Definition}

\begin{document}
\theoremstyle{remark}
\newtheorem{remark}{Remark}

%style file for ESANN manuscripts
\title{Model Agnostic Local Explanations of Reject}

%***********************************************************************
% AUTHORS INFORMATION AREA
%***********************************************************************
\author{Andr\'e Artelt\footnote{Corresponding author: \href{mailto:aartelt@techfak.uni-bielefeld.de}{aartelt@techfak.uni-bielefeld.de}}\;\,\footnote{Affiliation with the University of Cyprus}\;, Roel Visser and Barbara Hammer
%
% Optional short acknowledgment: remove next line if non-needed
\thanks{We gratefully acknowledge fundings from the Deutsche Forschungsgemeinschaft (DFG, German Research
Foundation) for grant TRR 318/1 2021 - 438445824, and the VW-Foundation for the project \textit{IMPACT} funded in the frame of the funding line \textit{AI and its Implications for Future Society}.}
%
% DO NOT MODIFY THE FOLLOWING '\vspace' ARGUMENT
\vspace{.3cm}\\
%
% Addresses and institutions (remove "1- " in case of a single institution)
CITEC -- Cognitive Interaction Technology \\
Bielefeld University -- Faculty of Technology \\
Inspiration 1, 33619 Bielefeld -- Germany
%
% Remove the next three lines in case of a single institution
%\vspace{.1cm}\\
%2- School of Second Author - Dept of Second Author \\
%Address of Second Author's school - Country of Second Author's school\\
}
%***********************************************************************
% END OF AUTHORS INFORMATION AREA
%***********************************************************************

\maketitle

\begin{abstract}
%Type your 100 words abstract here.
The application of machine learning based decision making systems in safety critical areas requires reliable high certainty predictions.

Reject options are a common way of ensuring a sufficiently high certainty of predictions made by the system. While being able to reject uncertain samples is important, it is also of importance to be able to explain why a particular sample was rejected. However, explaining general reject options is still an open problem.

We propose a model agnostic method for locally explaining arbitrary reject options by means of interpretable models and counterfactual explanations.
\end{abstract}

\section{Introduction}\label{sec:introduction}
Nowadays, machine learning (ML) based decision making system are omnipresent -- in particular, they are used in safety critical scenarios such as autonomous driving~\cite{AutonmousDriving}, credit (risk) assessment~\cite{CreditRiskML} and predictive policing~\cite{PredictivePolicing}. Trust and reliability are critical aspects of such decision making systems.

Trust can be realized by transparency -- i.e. it is difficult to trust a system that we do no not understand. It is common to achieve transparency by means of explanations -- i.e. providing explanations of the systems behavior~\cite{molnar2019}. There exist different explanation methods~\cite{molnar2019} such as feature relevance/importance methods and examples based methods such as contrasting explanations.

Reliability means that we require the system to consistently output high quality predictions. However, because the models are build to output a prediction for every possible input (no matter how plausible or implausible this might be), a high quality prediction can not always be guaranteed. In particular, the certainty of the prediction might vary a lot between different inputs. Uncertain predictions are problematic in scenarios where making mistakes can have serious consequences -- in such cases it might be better to refuse a prediction instead of making a potentially wrong prediction.
For instance consider the example of a spam and phishing mail filter: \textit{Imagine a mail filter application that tries to filter our spam and phishing mails in order to protect the end users and their surrounding from serious consequences. The filter is supposed to automatically sort out mails where it it is certain that the particular mails are malicious, and pass all benign mails to the user without any delay. However, in cases where the filter is not absolute certain about its prediction (distinguishing benign vs. malicious), it should reject this mail and pass it to a human for manually checking its content -- rejected mails might be passed to the user with an additional warning of taking care or to the it-security department of the company for further investigations and improvement of the filtering application. In order to understand the rejection and to support the further development of the filtering application, it is helpful to get an explanation why the filter was not able to classify the given mail.}

\paragraph*{Related Work and our Contributions}
Surprisingly, there does not exist a lot of work on explaining reject options. The only work we are aware of~\cite{explain_reject_lvq}, which deals with reject options for learning vector quantization (LVQ) models. However, their proposed method is completely tailored towards LVQ models and its specific reject options -- i.e. it is not applicable to any other models or reject options.

In this work, we propose a model agnostic method for locally explaining any reject option -- i.e. we propose a method that is applicable to any model and every possible reject option. Instead of globally explaining the given reject option, we aim for a local explanation -- i.e. explaining why a particular sample was rejected or not.

The remainder of this work is structured as follows: We first briefly review the necessary foundations in Section~\ref{sec:foundations} and then propose our model agnostic local explanation of reject options in Section~\ref{sec:modeling}. Subsequently, we empirically evaluate our proposed methods from Section~\ref{sec:modeling} in Section~\ref{sec:experiments}. The work finishes with a summary and conclusion in Section~\ref{sec:conclusion}.

\section{Foundations}\label{sec:foundations}
In the following, we briefly review the necessary foundations of this work. First, we introduce the general modeling of reject options in Section~\ref{sec:foundations:rejectoptions} and subsequently conformal prediction as a potential implementation of a reject option. Then, we briefly touch upon eXplanaible AI and discuss local approximations (see Section~\ref{sec:foundations:xai:localapprox}) and counterfactual explanations (see Section~\ref{sec:foundations:xai:counterfactuals}).

\subsection{Reject Options}\label{sec:foundations:rejectoptions}
Given an arbitrary classifier $\classifier:\setX\to\setY$, a reject option~\cite{hendrickx2021machine} is usually added by providing an additional function $\reject_{\classifier}:\setX\to\RN_{+}$ that measures the certainty of classifying $\x$ and rejects a sample $\x$ if the certainty is below a given threshold $\threshold$:
\begin{equation}\label{eq:rejectoption}
	\reject_{\classifier}(\x) < \threshold
\end{equation}
where the subscript $\classifier$ denotes a potential dependency on the classifier $\classifier(\cdot)$.

We can think about enriching a classifier $\classifier(\cdot)$ with a reject option as constructing a new classifier $\classifier':\setX\to\setY\cup\{\rejectSymbol\}$ where we add a reject symbol $\rejectSymbol$ to the set of possible predictions $\setY$:
\begin{equation}\label{eq:classifier_with_rejectoption}
    \classifier'(\x) =
    \begin{cases}
    \classifier(\x)       & \quad \text{if } \reject_{\classifier}(\x) \geq \threshold\\
    \rejectSymbol  & \quad \text{otherwise}
  \end{cases}
\end{equation}
In the following, we briefly introduce conformal prediction as a specific way of realizing such a reject option $\reject(\cdot)$. 

\subsubsection{Conformal Prediction for Implementing a Reject Option}
Assume that a (black-box) probabilistic classifier $\classifier:\setX\to\setY$ of the following form is given:
\begin{equation}\label{eq:classifierproba}
    \classifier(\x) = \underset{\y\,\in\,\setY}{\arg\max}\;p(\y \mid \x)
\end{equation}
where $p(\y \mid \x)$ denotes the class wise probability as estimated by the classifier $\classifier(\cdot)$.

A central building block of a conformal predictor~\cite{ConformalShaferV08} is a so called non-conformity measure $\nonconformity_{\classifier}:\setX,\setY\to\RN$ which measures how different a given labeled sample is from a given set of labeled samples we have seen before.
In case of a probabilistic classifier $\classifier(\cdot)$, a common non-conformity measure is given as follows:
\begin{equation}\label{eq:conformal:nonconformitymeasure}
    \nonconformity_{\classifier}(\x,\y=j) = \underset{i\neq j}{\max}\; p_{\classifier}(\y=i\mid\x) - p_{\classifier}(\y=j\mid\x)    
\end{equation}

For calibrating (fitting) a conformal predictor based on $\classifier(\cdot)$, we need another labeled data set $\Dcalib\subset\setX\times\setY$ which was not used during the fitting of $\classifier(\cdot)$.
Calibrating/fitting an (inductive) conformal predictor means to compute the non-conformity $\alpha_i$ of every sample from the calibration set by applying $\nonconformity_{\classifier}(\cdot)$:
\begin{equation}\label{eq:conformal:calibration}
    \alpha_i = \nonconformity_{\classifier}(\x_i, \y=\y_i)
\end{equation}
For every new data point $\xnew\in\setX$ that is going to be classified, we compute the non-conformity measure for every possible label in $\setY$:
\begin{equation}
    \alpha_{*}^i = \nonconformity_{\classifier}(\xnew, \y=i)
\end{equation}
Next, the non-conformity scores of $\xnew$ are compared with the non-conformity scores from the calibration set to compute p-values for every possible classification of $\xnew$:
\begin{equation}\label{eq:conformal:pvalue}
    p_{\y=i}(\xnew) =\frac{|\alpha_j\in\Dcalib \geq \alpha_{*}^i|}{|\Dcalib|+1} 
\end{equation}
The conformal predictor then selects the label with the larges p-value as a prediction -- i.e.~\refeq{eq:classifierproba} becomes:
\begin{equation}
    \classifier(\xnew) = \underset{i\,\in\,\setY}{\arg\max}\;p_{\y=i}(\xnew)
\end{equation}
The confidence of the prediction -- i.e. how likely (given the training set) the prediction is compared to all other possible predictions -- is then computed as follows:
\begin{equation}\label{eq:conformal:confidence}
    1 - \left(\underset{j\neq \underset{i}{\argmax}\,p_{\y=i}(\xnew)}{\max}\;p_{\y=j}(\xnew)\right)
\end{equation}
and the credibility -- i.e. how well the training set supports the prediction -- is given as follows:
\begin{equation}\label{eq:conformal:credibility}
    \credibility(\xnew) = \underset{i}{\max}\;p_{\y=i}(\xnew)
\end{equation}

In order to use conformal prediction for implementing a reject option~\refeq{eq:rejectoption}~\cite{linusson2018classification}, one could use either the conformity score~\refeq{eq:conformal:confidence} or the credibility score~\refeq{eq:conformal:credibility}. As it is common practice, we use the credibility as a reject score in this work:
\begin{equation}\label{eq:reject:credibility}
    \reject_{\classifier}(\x) = \credibility(\x)
\end{equation}

\subsection{Explanations}
In the following, we briefly review two popular types of explanations. Explanations using local approximations (Section~\ref{sec:foundations:xai:localapprox}) and counterfactual explanations as an instance of example based explanations (Section~\ref{sec:foundations:xai:counterfactuals}).

\subsubsection{Local Approximations}\label{sec:foundations:xai:localapprox}
There exist popular methods for locally explaining a given model $\classifier(\cdot)$, instead of trying to come up with a global explanation~\cite{molnar2019}. A common approach for local explanations is to build a local approximation of the model $\classifier(\cdot)$ which is then used for creating an explanation.

A popular instance of such methods is LIME~\cite{DBLP:conf/kdd/Ribeiro0G16}. Here, the authors propose to fit an interpretable model (e.g. a linear model) to a set of labeled perturbed samples -- i.e. the original model  $\classifier(\cdot)$ is applied to perturbed instances of the given original sample $\xorig$ for which we want to compute a local explanation. The final, local, explanation is then constructed using the most relevant features of the local approximation -- in order to get a meaningful explanation, the features must be interpretable and meaningful (e.g. super-pixels in case of images).

Another method that uses local approximations for computing a local explanation is Anchors~\cite{DBLP:conf/aaai/Ribeiro0G18}. Anchors are if-then rules based explanations that locally explain the prediction of the given model $\classifier(\cdot)$.

\subsubsection{Counterfactual Explanations}\label{sec:foundations:xai:counterfactuals}
Counterfactual explanations (often just called \textit{counterfactuals}) are a prominent instance of contrasting explanations, which state a change to some features of a given input such that the resulting data point, called the counterfactual, causes a different behavior of the system than the original input does. Thus, one can think of a counterfactual explanation as a suggestion of actions that change the model's behavior/prediction. One reason why counterfactual explanations are so popular is that there exists evidence that explanations used by humans are often contrasting in nature~\cite{CounterfactualsHumanReasoning} -- i.e. people often ask questions like \textit{``What would have to be different in order to observe a different outcome?''}.
%For illustrative purposes, consider the example of loan application: \textit{Imagine you applied for a credit at a bank. Unfortunately, the bank rejects your application. Now, you would like to know why. In particular, you would like to know what would have to be different so that your application would have been accepted.
%A possible explanation might be that you would have been accepted if you had earned 500\$ more per month and if you hand not had a second credit card.}
Despite their popularity, the missing uniqueness of counterfactuals could pose a problem: Often there exist more than one possible/valid counterfactual -- this is called the Rashomon effect~\cite{molnar2019} -- and in such cases, it is not clear which or how many of them should be presented to the user. One common modeling approach (if this problem is not simply ignored) is to enforce uniqueness by a suitable formalization.

In order to keep the explanation (suggested changes) simple -- i.e. easy to understand -- an obvious strategy is to look for a small number of changes so that the resulting sample (counterfactual) is similar/close to the original sample, which is aimed to be captured by~\refdef{def:counterfactual}.
\begin{definition}[(Closest) Counterfactual Explanation~\cite{CounterfactualWachter}]\label{def:counterfactual}
Assume a prediction function $\classifier:\RN^\dimsym \to \setY$ is given. Computing a counterfactual $\xcf \in \RN^\dimsym$ for a given input $\xorig \in \RN^\dimsym$ is phrased as an optimization problem:
\begin{equation}\label{eq:counterfactualoptproblem}
\underset{\xcf \,\in\, \RN^\dimsym}{\arg\min}\; \loss\big(\classifier(\xcf), \ycf\big) + C \cdot \regularization(\xcf, \xorig)
\end{equation}
where $\loss(\cdot)$ denotes a loss function, $\ycf$ the target prediction, $\regularization(\cdot)$ a penalty for dissimilarity of $\xcf$ and $\xorig$, and $C>0$ denotes the regularization strength.
\end{definition}
In the following, we assume a binary classification problem: In this case, we denote a (closest) counterfactual $\xcf$ according to~\refdef{def:counterfactual} of a given sample $\xorig$ under a prediction function $\classifier(\cdot)$ simply as $
\xcf=\myCF{\xorig}{\classifier}$ and drop the target label $\ycf$ because it is uniquely determined.

The counterfactuals from~\refdef{def:counterfactual} are also called \textit{closest counterfactuals} because the optimization problem~\refeq{eq:counterfactualoptproblem} tries to find an explanation $\xcf$ that is as close as possible to the original sample $\xorig$. However, other aspects like plausibility and actionability are ignored in~\refdef{def:counterfactual}, but are covered in other work~\cite{CounterfactualGuidedByPrototypes,PlausibleCounterfactuals,ActionableCounterfactuals} -- note that it is not always clear which type of counterfactual is meant when people talk about counterfactuals. In this work, we use the term counterfactuals in the spirit of~\refdef{def:counterfactual}.

\section{Local Approximations for Explaining Reject}\label{sec:modeling}
We propose a model agnostic approach for locally explaining arbitrary reject options -- i.e. our method does not need access to the reject option or the underlying ML model, access to a prediction interface is sufficient. Instead of explaining the reject option globally, we aim for a local explanation only -- i.e. explaining the reject of a particular sample.

Given a sample $\xorig\in\setX$ which is rejected by the reject option, we sample a fixed number of samples $\{\x_i\}$ from the neighborhood around $\xorig$ and label each sample whether it is also rejected or not:
\begin{equation}
    \y_i = \begin{cases}
        1 & \quad \text{if } \reject(\x_i) < \threshold \\
        0 & \quad \text{otherwise}
    \end{cases} \quad\quad\quad \forall\,\x_i \in \set{B}_{\epsilon}(\xorig)
\end{equation}
where $\set{B}_{\epsilon}(\xorig)$ denotes a fixed number of samples in the neighborhood of $\xorig$.
We then fit an interpretable classifier $\classifierLocal$ (e.g. a linear model or a decision tree) to these samples $\Dlocal=\{(\x_i, \y_i)\}$.

We propose to either use $\classifierLocal(\cdot)$ as an explanation -- e.g. using the obtained feature importances or learned decision rules as an explanation --, or a counterfactual explanation (see~\refdef{def:counterfactual}) $\xcf=\myCF{\xorig}{\classifierLocal}$ of $\classifierLocal(\cdot)$ as an explanation of the reject of $\xorig$.

Formally, we propose two different realizations of a local explanation $\explanation$ at $\xorig$ under a given reject option $\reject(\cdot)$:
\begin{equation}
    \explanation(\reject, \xorig) = \begin{cases}
        \featurerelvance(\classifierLocal) \\
        \myCF{\xorig}{\classifierLocal} 
    \end{cases}
\end{equation}%TODO: Do some theory for these types of explanations
where $\featurerelvance(\cdot)$ denotes the feature relevance as obtained from a given model.
In the experiments (Section~\ref{sec:experiments}), we empirically evaluate and compare both types of explanations in the experiments (Section~\ref{sec:experiments}).

\section{Experiments}\label{sec:experiments}
In the following, we empirically evaluate our proposed model agnostic methods for explaining rejects (see Section~\ref{sec:modeling}).
We do so by considering two different aspects for evaluation:
\begin{itemize}
    \item We evaluate computational aspects like sparsity of the computed explanations.
    \item We evaluate the ground truth recovery rate (goodness) of the explanations by evaluating if and how well the explanations match the ground truth -- i.e. identifying the relevant features.
\end{itemize}
In addition, we always evaluate the accuracy of the learned local approximation -- i.e. checking if the original sample is also rejected under the local approximation.
All experiments are implemented in Python and the implementation is publicly available on GitHub\footnote{\url{https://github.com/andreArtelt/LocalModelAgnosticExplanationReject}}.

\subsection{Data Sets}
We consider the following data sets for our empirical evaluation -- all data sets are scaled and standardized:

\subsubsection{Wine}
The ``Wine data set''~\cite{winedata} is used for predicting the cultivator of given wine samples based on their chemical properties. The data set contains $178$ samples and $13$ numerical features such as alcohol and color intensity.

\subsubsection{Breast cancer}
The ``Breast Cancer Wisconsin (Diagnostic) Data Set''~\cite{breastcancer} is used for classifying breast cancer samples into benign and malignant. The data set contains $569$ samples and $30$ numerical features such as smoothness and compactness.

\subsubsection{Flip}
This data set~\cite{flipdata} is used for the prediction of fibrosis. The set consists of samples of $118$ patients and $12$ numerical features such as blood glucose, BMI and total cholesterol.
As the data set contains some rows with missing values, we chose to replace these missing values with the corresponding feature mean.

\subsubsection{t21}
This data set~\cite{t21dataset} is used for early diagnosis of chromosomal abnormalities, such as trisomy 21, in pregnant women. The data set consists of $18$ numerical features such as heart rate and weight, and contains over $50000$ samples but only $0.8$ percent abnormal samples (e.g. cases of trisomy 21) -- i.e. it is highly imbalanced.

\subsection{Setup}
Since our method (see Section~\ref{sec:modeling}) is completely model agnostic, we evaluate it on a set of diverse classifiers: k-nearest neighbors classifier (kNN), Gaussian naive Bayes classifier (GNB), random forest classifier (RandomForest).
Whereby we always use conformal prediction (see Section~\ref{sec:foundations:rejectoptions}) for realizing a credibility based reject option~\refeq{eq:reject:credibility}.

In order to make a fair comparison between the efficacy of the methods we perform hyperparameter tuning in order to find the best performing model parameters. This includes the hyperparameters of the respective classifiers, which are obtained by a grid search on each of them. Additionally, we try to find an appropriate rejection threshold by using the Knee/Elbow method \cite{satopaa2011kneedle} for finding a reasonable cut-off point. Using the Kneedle algorithm~\cite{satopaa2011kneedle}, we can find an appropriate rejection threshold by determining the ``optimal'' threshold in the ARC by finding the so-called knee-point.
In a real world scenario the threshold might be tuned to allow for a more relaxed or strict rejection scenario, however for the purpose of our research finding the knee point gives us a fairly good approximation of what would usually be considered an appropriate or well performing rejection threshold.

We run all experiments in a $5$-fold cross validation to get meaningful and statistically reliable results -- we consider every possible combination of data set and classifier. We use a decision tree classifier for the computation of the local approximation.
After fitting the classifier, we apply the reject option to all samples from the test set and compute explanations for those that are rejected by the reject option. As proposed in Section~\ref{sec:modeling}, we always compute two explanations: feature relevance profile as obtained from the local approximation -- we use the Gini importance as obtained from the decision tree classifier -- and a counterfactual explanation under this local approximation.

\paragraph*{Algorithmic Properties}
When evaluating algorithmic properties, we not only compute the accuracy -- i.e. is the prediction of the local approximation consistent with the prediction of the original model --, but also compute the sparsity ($\lZero$-norm) of both explanations.

\paragraph*{Goodness of Explanations}
For evaluating the goodness of the explanations, we create scenarios with known ground truth as follows: For each data set, we select a random subset of features ($30\%$) and perturb these in the test set by adding Gaussian noise -- we then check which of these samples are rejected due to the noise (i.e. applying the reject option before and after applying the perturbation), and compute explanations of these samples only.
Finally, we evaluate for both explanations how many of the relevant features (from the known ground truth) are recovered and included in the explanation.

\subsection{Results \& Discussion}
When reporting the results, we use the following abbreviations: \textit{FeatImp} -- Feature importances as obtained from the local approximation; \textit{Cf} -- Counterfatual explanation.
Note that we round all values to two decimal points.

\paragraph*{Algorithmic Properties}
We report the mean accuracy and sparsity in Table~\ref{table:experimentresults:algoproperties}.
\begin{table}
\caption{Algorithmic properties -- Mean (incl. variance) accuracy and sparsity -- larger values are ``better'' for accuracy, while smaller values are ``better'' for sparsity.}
\centering
\footnotesize
\begin{tabular}{|c|c||c||c|c||}
 \hline
 & \textit{DataSet} & Accuracy & FeatImp & Cf\\
 \hline
 \multirow{5}{*}{\rotatebox[origin=l]{0}{kNN}}
 & Wine & $0.80 \pm 0.16$ & $4.5 \pm 1.98$ & $1.25 \pm 0.23$ \\
 & Breast Cancer & $0.92 \pm 0.0$ & $5.12 \pm 1.66$ & $1.25 \pm 0.19$ \\
 & t21 & $0.96 \pm 0.00$ & $3.9 \pm 3.43$ & $1.07 \pm 0.27$ \\
 & Flip & $0.31 \pm 0.07$ & $5.21 \pm 1.13$ & $1.00 \pm 0.00$ \\
 \hline\hline
 \multirow{5}{*}{\rotatebox[origin=l]{0}{GNB}}
 & Wine & $0.92 \pm 0.00$ & $4.57 \pm 1.17$ & $1.11 \pm 0.10$ \\
 & Breast Cancer & $0.88 \pm 0.00$ & $3.83 \pm 1.38$ & $1.07 \pm 0.07$ \\
 & t21 & $0.78 \pm 0.15$ & $1.12 \pm 1.71$ & $0.71 \pm 0.26$ \\
 & Flip & $0.83 \pm 0.01$ & $1.73 \pm 0.54$ & $1.00 \pm 0.00$ \\
 \hline\hline
 \multirow{5}{*}{\rotatebox[origin=l]{0}{RandomForest}}
 & Wine & $0.8 \pm 0.16$ & $3.26 \pm 1.64$ & $1.43 \pm 0.37$ \\
 & Breast Cancer & $1.00 \pm 0.00$ & $1.07 \pm 2.35$ & $0.52 \pm 0.48$ \\
 & t21 & $0.95 \pm 0.00$ & $3.75 \pm 2.59$ & $1.22 \pm 0.30$ \\
 & Flip & $0.50 \pm 0.06$ & $5.05 \pm 1.33$ & $1.05 \pm 0.05$ \\
 \hline
\end{tabular}
\label{table:experimentresults:algoproperties}
\end{table}
We observe that the local approximation is usually sufficiently good (although some combinations of model and data set seem to be more challenging) and the final explanations are very sparse -- i.e. we obtain low-complexity explanations. Furthermore, we observe that counterfactual explanations of the local approximation are consistently sparser than the obtained feature importance.

\paragraph*{Goodness of Explanations}
The mean recall of correctly recovered relevant features is given in Table~\ref{table:experimentresults:goodness}.
\begin{table}
\caption{Goodness of explanations -- Mean (incl. variance) recall of correctly identified relevant features (larger numbers are better).}
\centering
\footnotesize
\begin{tabular}{|c|c||c||c|c||}
 \hline
 & \textit{DataSet} & Accuracy & FeatImp & Cf \\
 \hline
 \multirow{5}{*}{\rotatebox[origin=l]{0}{kNN}}
 & Wine & $0.75 \pm 0.15$ & $0.53 \pm 0.03$ & $0.28 \pm 0.15$\\
 & Breast Cancer & $0.89 \pm 0.02$ & $0.50 \pm 0.04$ & $0.23 \pm 0.12$ \\
 & t21 & $0.78 \pm 0.02$ & $0.56 \pm 0.03$ & $0.36 \pm 0.15$ \\
 & Flip & $0.40 \pm 0.14$ & $0.30 \pm 0.08$ & $0.04 \pm 0.03$ \\
 \hline\hline
 \multirow{5}{*}{\rotatebox[origin=l]{0}{GNB}}
 & Wine & $0.85 \pm 0.04$ & $0.56 \pm 0.06$ & $0.43 \pm 0.18$ \\
 & Breast Cancer & $0.97 \pm 0.0$ & $0.39 \pm 0.09$ & $0.23 \pm 0.12$ \\
 & t21 & $0.60 \pm 0.24$ & $0.45 \pm 0.13$ & $0.36 \pm 0.15$ \\
 & Flip & $0.91 \pm 0.01$ & $0.40 \pm 0.14$ & $0.38 \pm 0.18$ \\
 \hline\hline
 \multirow{5}{*}{\rotatebox[origin=l]{0}{RandomForest}}
 & Wine & $1.00 \pm 0.00$ & $0.51 \pm 0.13$ & $0.39 \pm 0.16$ \\
 & Breast Cancer & $1.00 \pm 0.00$ & $0.18 \pm 0.08$ & $0.16 \pm 0.09$ \\
 & t21 & $0.62 \pm 0.11$ & $0.58 \pm 0.05$ & $0.50 \pm 0.15$ \\
 & Flip & $0.61 \pm 0.08$ & $0.54 \pm 0.08$ & $0.38 \pm 0.15$ \\
 \hline
\end{tabular}
\label{table:experimentresults:goodness}
\end{table}
First, we observe that the perturbation does not strongly affect the accuracy. Next, we observe that both explanations have trouble to recover all perturbed features -- although the feature importance explanation recovers consistently more perturbed features than the counterfactual explanations. The reasons for this are two-fold: First, because we optimized sparsity (i.e. getting low-complexity explanations), the explanations contain very few features only and are therefore likely to miss some perturbed features. Second, it seems that the local approximation is not sensitive enough to the applied perturbations -- the accuracy is pretty high, but still the explanations have trouble identifying all perturbed features.

\section{Summary \& Conclusion}\label{sec:conclusion}
In this work, we proposed a model agnostic approach for explaining reject options: we proposed to use a local approximation of the reject option and explain the reject locally either by the local approximation itself (assuming that this local approximation is interpretable) or by counterfactual explanations of this local approximation.
We empirically evaluated these two explanations methods under computational as well as qualitative aspects. We observed reasonable performance of both explanations -- in particular counterfactual explanations were able to come up with low complexity explanations but identified fewer of the relevant features.

The empirical evaluation in this work focuses on computational proxies only. However, it still remains unclear if and how useful our proposed explanations are to humans. Since it is difficult to phrase ``usefullness'' as a scoring function, a proper use study is needed.
We leave this aspects as future work.

% ****************************************************************************
% BIBLIOGRAPHY AREA
% ****************************************************************************

\begin{footnotesize}

% IF YOU USE BIBTEX,
% - DELETE THE TEXT BETWEEN THE TWO ABOVE DASHED LINES
% - UNCOMMENT THE NEXT TWO LINES AND REPLACE 'Name_Of_Your_BibFile'

\bibliographystyle{unsrt}
\bibliography{bibliography}

\end{footnotesize}

% ****************************************************************************
% END OF BIBLIOGRAPHY AREA
% ****************************************************************************

\end{document}